\documentclass[letterpaper, 10 pt, conference]{ieeeconf}  

\IEEEoverridecommandlockouts                
\usepackage{graphicx}
\usepackage{hyperref}
\usepackage{booktabs}
\usepackage[ruled,vlined,linesnumbered,boxed,noend]{algorithm2e}
\usepackage{amsmath, amssymb}
\usepackage{balance}
\usepackage{tikz}
\usetikzlibrary{arrows,automata, positioning, calc}
\usepackage{cleveref}
\usepackage{soul}
\usepackage{siunitx}

\title{\LARGE \bf
Dynamic Interaction-Aware Scene Understanding for Reinforcement Learning in Autonomous Driving
}

\author{
  Maria Huegle$^{1}$, Gabriel Kalweit$^{1}$, Moritz Werling$^{2}$ and Joschka Boedecker$^{1,3}$\\
\thanks{$^{1,3}$Dept. of Computer Science, University of Freiburg, Germany.}%
\thanks{{\tt \{hueglem,kalweitg,jboedeck\}@cs.uni-freiburg.de}}
\thanks{$^{2}$BMWGroup, Unterschleissheim, Germany.}%
\thanks{{\tt  Moritz.Werling@bmw.de}}%
\thanks{$^{3}$Cluster of Excellence BrainLinks-BrainTools, Freiburg, Germany.}
}

\begin{document}

\maketitle
\thispagestyle{empty}
\pagestyle{empty}

\begin{abstract}

The common pipeline in autonomous driving systems is highly modular and includes a perception component which extracts lists of surrounding objects and passes these lists to a high-level decision component. In this case, leveraging the benefits of deep reinforcement learning for high-level decision making requires special architectures to deal with multiple variable-length sequences of different object types, such as vehicles, lanes or traffic signs. At the same time, the architecture has to be able to cover interactions between traffic participants in order to find the optimal action to be taken. In this work, we propose the novel Deep Scenes architecture, that can learn complex interaction-aware scene representations based on extensions of either 1) Deep Sets or 2) Graph Convolutional Networks. We present the Graph-Q and DeepScene-Q off-policy reinforcement learning algorithms, both outperforming state-of-the-art methods in evaluations with the publicly available traffic simulator SUMO. 
\end{abstract}

\section{INTRODUCTION}

 In autonomous driving scenarios, the number of traffic participants and lanes surrounding the agent can vary considerably over time. Common autonomous driving systems use modular pipelines, where a perception component extracts a list of surrounding objects and passes this list to other modules, including localization, mapping, motion planning and high-level decision making components. Classical rule-based decision-making systems are able to deal with variable-sized object lists, but are limited in terms of generalization to unseen situations or are unable to cover all interactions in dense traffic. Since Deep Reinforcement Learning (DRL) methods can learn decision policies from data and off-policy  methods can improve from previous experience, they offer a promising alternative to rule-based systems. In the past years, DRL has shown promising results in various domains \cite{DBLP:journals/nature/MnihKSRVBGRFOPB15, DBLP:journals/nature/SilverHMGSDSAPL16, DBLP:conf/nips/WatterSBR15, DBLP:journals/jmlr/LevineFDA16, Mirchevska2018HighlevelDM}.  However, classical DRL architectures like fully-connected or convolutional neural networks (CNNs) are limited in their ability to  deal with variable-sized, structured inputs or to model interactions between objects. 
 
 Prior works on reinforcement learning for autonomous driving that used fully-connected network architectures and fixed sized inputs \cite{adaptive_behavior_kit, branka_rl_highway, Mirchevska2018HighlevelDM, 2018TowardsPH, overtaking_maneuvers_kaushik} are limited in the number of vehicles that can be considered. CNNs using occupancy grids  \cite{tactical_decision_making, deeptraffic} are limited to their initial grid size. Recurrent neural networks are useful to cover temporal context, but are not able to handle a variable number of objects permutation-invariant  w.r.t to the input order for a fixed time step. In \cite{DeepSetQ}, limitations of these architectures are shown and a more flexible architecture based on Deep Sets \cite{NIPS2017_6931} is proposed for off-policy reinforcement learning of lane-change maneuvers, outperforming traditional approaches in evaluations with the open-source simulator SUMO.
 
In this paper, we propose to use Graph Networks \cite{DBLP:journals/corr/KipfW16} as an interaction-aware input module in reinforcement learning for autonomous driving. We employ the structure of Graphs in off-policy DRL and formalize the Graph-Q algorithm. In addition, to cope with multiple object classes of different feature representations, such as different vehicle types, traffic signs or lanes, we introduce the formalism of Deep Scenes, that can extend  Deep Sets and Graph Networks to fuse multiple variable-sized input sets of different feature representations. Both of these can be used in our novel DeepScene-Q algorithm for off-policy DRL. Our main contributions are:
\begin{enumerate}
    \item Using Graph Convolutional Networks to model interactions between vehicles in DRL for autonomous driving.
    \item Extending existing set input architectures for DRL to deal with multiple lists of different object types.
\end{enumerate}

\section{RELATED WORK}

Graph Networks are a class of neural networks that can learn functions on graphs as input \cite{Scarselli:2009:GNN:1657477.1657482, DBLP:journals/corr/abs-1812-08434, DBLP:journals/corr/abs-1806-01242,DBLP:journals/corr/BattagliaPLRK16, DBLP:journals/corr/abs-1806-01261} and can reason about how objects in complex systems interact. They can be used in DRL to learn state representations \cite{DBLP:journals/corr/DaiKZDS17, DBLP:journals/corr/abs-1806-01203, DBLP:journals/corr/abs-1810-09202, DBLP:journals/corr/abs-1806-01242}, e.g. for inference and control of physical systems with bodies (objects) and joints (relations). In the application for autonomous driving, Graph Networks were used for supervised traffic prediction while modeling traffic participant interactions \cite{DBLP:journals/corr/abs-1903-01254},  where vehicles were modeled as objects and interactions between them as relations. Another type of interaction-aware network architectures, Interaction Networks, were proposed to reason about how objects in complex systems interact  \cite{DBLP:journals/corr/BattagliaPLRK16}. A vehicle behavior interaction network that captures vehicle  interactions was presented in \cite{DBLP:journals/corr/abs-1903-00848}. In \cite{DBLP:journals/corr/abs-1805-06771}, a convolutional social pooling component was proposed using a CNN to model spatial connections between vehicles for vehicle trajectory prediction.

  \begin{figure*}[h]
    \centering
    (a) \hspace{-1mm} \includegraphics[width=0.30\textwidth]{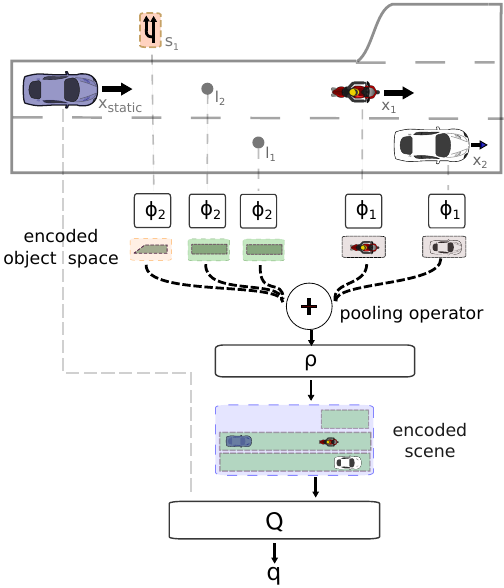}
    \hspace{10mm} (b) \includegraphics[width=0.31\textwidth]{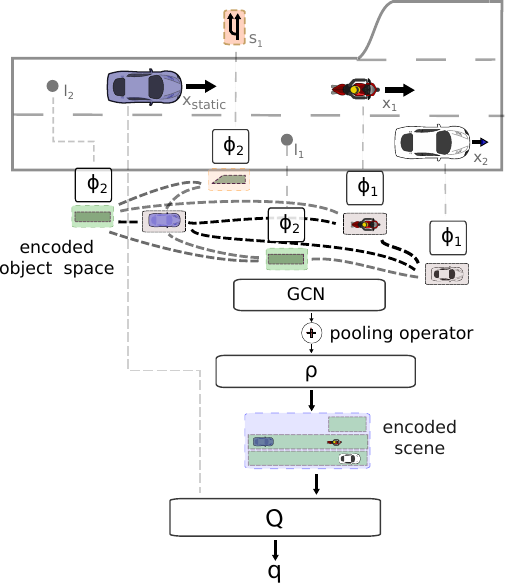}
\caption{Scheme of DeepScene-Q, using (a) Deep Sets and (b) Graphs. Both architectures combine multiple variable-length object lists in a scene, here a traffic sign $s_1$, lanes $l_{1}, l_2$ and vehicles $x_1, x_2$. The modules $\phi_i$, $\rho$ and $Q$ are fully-connected networks. As permutation invariant pooling operator, we use the \textit{sum}. The vector $x^{\text{static}}$ includes static features and $q$ the action value output.}
\label{fig:deepscene}
\end{figure*}

\section{PRELIMINARIES}

We model the task of high-level decision making for autonomous driving as a Markov Decision Process (MDP), where the agent is following a policy $\pi$ in an environment in a state $s_t$, applying a discrete action $a_t \sim \pi$ to reach a successor state $s_{t+1}\sim\mathcal{M}$ according to a transition model $\mathcal{M}$. In every time step $t$, the agent receives a reward $r_{t}$, e.g. for driving as close as possible to a desired velocity. The agent tries to maximize the discounted long-term return $R(s_t) = \sum_{i>=t} \gamma^{i-t}r_i$, where $\gamma \in [0, 1]$ is the discount factor. In this work, we use Q-learning \cite{Watkins92q-learning}. The Q-function $Q^\pi(s_t,a_t)=\mathbf{E}_{a_{i>t}\sim\pi}[R(s_t)|a_t]$ represents the value of following a policy $\pi$  after applying action $a_t$. The optimal policy can be inferred from the optimal action-value function $Q^*$ by maximization over actions.

\subsection{Q-Function Approximation}

 We use DQN \cite{DBLP:journals/nature/MnihKSRVBGRFOPB15} to  estimate the optimal $Q$-function by function approximator $Q$, parameterized by $\theta^{Q}$. It is trained in an offline fashion on minibatches sampled from a fixed replay buffer $\mathcal{R}$ with transitions collected by a driver policy $\hat{\pi}$. As loss, we use
$L(\theta^{Q}) = \frac{1}{b}\sum_i \left(y_i - Q(s_i, a_i| \theta^{Q})\right)^2$ with targets $y_i=r_i + \gamma \max_{a} Q'(s_{i+1}, a| \theta^{Q'}),$ where $Q'$ is a target network, parameterized by $\theta^{Q'}$, and  $(s_i, a_i, s_{i+1}, r_i)|_{0\leq i \leq b}$ is a randomly sampled minibatch from $\mathcal{R}$.  For the target network, we use a soft update, i.e. $\theta^{Q'}\leftarrow\tau\theta^{Q}+ (1-\tau)\theta^{Q'}$ with update step-size $\tau \in [0,1]$. Further, we use a variant of Double-$Q$-learning \cite{DBLP:journals/corr/HasseltGS15} which is based on two \textit{Q}-network pairs and uses the minimum of the predictions for the target calculation, similar as in \cite{DBLP:conf/icml/FujimotoHM18}.

\subsection{Deep Sets}

A network $Q_{\mathcal{DS}}$ can be trained to estimate the $Q$-function for a state representation $s=(X^{\text{dyn}}, x^{\text{static}})$ and action $a$. The representation consists of a static input $x^{\text{static}}$ and a dynamic, variable-length input set $X^{\text{dyn}}=[x^1, .., x^{\text{seq len}}]^\top$, where  $x^j|_{1 \le j \le \text{seq len}}$  are feature vectors for surrounding vehicles in sensor range.  In \cite{DeepSetQ}, it was proposed to use Deep Sets to handle this input representation, where the Q-network consists of three network modules $\phi,\rho$ and $Q$. The representation of the dynamic input set is computed by $ \Psi(X^{\text{dyn}})= \rho\left(\sum_{x\in X^{\text{dyn}}}\phi(x) \right),$
 which makes the Q-function permutation invariant w.r.t. the order of the dynamic input \cite{NIPS2017_6931}. Static feature representations $x^{\text{static}}$ are fed directly to the $Q$-module, and the Q-values can be computed by $Q_{\mathcal{DS}} = Q(\Psi(X^{\text{dyn}}) || x^{\text{static}})$, where $||$ denotes a concatenation of two vectors.  The Q-learning algorithm is called DeepSet-Q \cite{DeepSetQ}.

\section{METHODS}
\label{sec:methods}

\subsection{Deep Scene-Sets}
\label{sec:deepscenes}

 To overcome the limitation of DeepSet-Q to one variable-sized list of the same object type, we propose a novel architecture, Deep Scene-Sets, that are able to deal with $K$ input sets $X^{\text{dyn}_1}, ..., X^{\text{dyn}_K}$, where every set has variable length. A combined, permutation invariant representation of all sets can be computed by 
$$ \Psi(X^{\text{dyn}_1}, ..., X^{\text{dyn}_K}) = \rho \left( \sum_{k} \sum_{x \in X^{ \text{dyn}_k}}\phi^k(x) \right),$$

where $1 \le k \le K$. The output vectors $\phi^k(\cdot) \in \mathbb{R}^F $ of the neural network modules $\phi^k$ have the same length $F$. We additionally propose to share the parameters of the last layer for the different $\phi$ networks. Then, $\phi^k(\cdot)$ can be seen as a projection of all input objects to the same encoded \textit{object space}. We combine the encoded objects of different types by the \textit{sum} (or other permutation invariant pooling operators, such as \textit{max}) and use the network module $\rho$ to create an encoded \textit{scene}, which is a fixed-sized vector. The encoded scene is concatenated to $x^{\text{static}}$ and the Q-values can be computed by $ Q_\mathcal{D} =  Q(\Psi(X^{\text{dyn}_1}, ..., X^{\text{dyn}_K}) || x^{\text{static}})$.    We call the corresponding Q-learning algorithm DeepScene-Q, shown in Algorithm \ref{alg:sceneq} (Option 1) and \Cref{fig:deepscene} (a).

\begin{algorithm}[t]
\small
    \SetAlgoLined
    \DontPrintSemicolon
    initialize $Q_{\mathcal{G}}=(\phi,\rho, H, Q)$ and $Q'_{\mathcal{G}}=(\phi',\rho', H', Q')$, set replay buffer $\mathcal{R}$\\
    \For{\text{optimization step} o=1,2,\dots}{
     \SetKwProg{Fn}{}{:}{}
    
        get minibatch $(s_i,a_i,(X^{\text{dyn}}_{i+1}, x_{i+1}^{\text{static}}),r_{i+1})$ from $\mathcal{R}$\\
        \ForEach{\text{transition}}{
          \ForEach{\text{object }$x_{i+1}^j$ in $X^{\text{dyn}}_{i+1}$}{
                $(\phi'_{i+1})^j=\phi'\left(x_{i+1}^j\right)$\\
        }
             compute $H^{'(L)}_{i+1}$ by GCN with $H^{'(0)}_{i+1} = [(\phi'_{i+1})^1, ..., (\phi'_{i+1})^\text{seq len}]^\top$\;
             get  $\rho'_{i+1}=  \rho'\left(\sum\limits_k \sum\limits_j H^{'(L)}_{i+1}\right)$\;
            }
            $y_i = r_{i+1}+\gamma\max_a Q'(\rho'_{i+1}, x_{i+1}^{\text{static}}, a)$\\
        
        perform a gradient step on loss: $\frac{1}{b}\sum\limits_i\left(Q_{\mathcal{G}}(s_i, a_i)-y_i\right)^2$\\
        update target network by: $\theta^{Q_{\mathcal{G}}'}\leftarrow\tau\theta^{Q_{\mathcal{G}}} + (1-\tau)\theta^{Q_{\mathcal{G}}'}$
    }
    \caption{Graph-Q}
    \label{alg:graphq}
\end{algorithm}

\subsection{Graphs}
In the Deep Set architecture, relations between vehicles are not explicitly modeled and have to be inferred in $\rho$. We extend this approach by using Graph Networks, considering graphs as input. Graph Convolutional Networks (GCNs) \cite{DBLP:journals/corr/KipfW16}  operate on graphs defined by a set of node features $X^{\text{dyn}}=[x^1, .., x^{\text{seq len}}]^\top$  and a set of edges represented by an adjacency matrix $A$. The propagation rule of the GCN is $H^{(l)}= \sigma(D^{\frac{1}{2}} \tilde A D^{\frac{1}{2}} H^{(l-1)}W^{(l-1)}) \text{ with } 1 \le l \le L,$ where we set $H^{(0)} = [\phi(x_1), ..., \phi(x_\text{seq len})]^\top$ using an encoder module similar as in the Deep Sets approach. $\tilde A \in \mathbb{R}^{N \times N}$ is an  adjacency matrix with added self-connections, $D_{i,i} = \sum_j \tilde A_{i,j}$, $\sigma$ the activation function, $H^{(l)} \in \mathbb{R}^{N \times F}$ hidden layer activations and $W^{(l)}$ the learnable  matrix of the $l$-th layer. The dynamic input representation can be computed from the last layer $L$ of the GCN: $ \Psi(X^{\text{dyn}})= \rho\left(\sum_{x \in X^{\text{dyn}}} H^{(L)} \right),$
 where $\phi$ is a neural network and the output vector $\phi(\cdot) \in \mathbb{R}^F $ has length $F$.  The Q-values can be computed by $ Q_\mathcal{G} =  Q(\Psi(X^{\text{dyn}}) || x^{\text{static}})$. We call the corresponding Q-learning algorithm  Graph-Q, see \Cref{alg:graphq}.

\subsection{Deep Scene-Graphs}

The graph representation can be extended to deal with multiple variable-length lists of different object types $X^{\text{dyn}_1}, ..., X^{\text{dyn}_K}$ by using $K$ encoder networks. As node features, we use $H^{(0)} = [\Phi^1, ..., \Phi^K]^\top$ and $\Phi^k = [\phi^k(x_1), ..., \phi^k(x_{\text{seq len}_k})]  \text{ for  } 1 \le k \le K,$
and compute the dynamic input representation from the last layer of the GCN:
 $$ \Psi(X^{\text{dyn}_1}, ..., X^{\text{dyn}_K})= \rho\left(\sum_{k} \sum_{x \in X^{ \text{dyn}_k}} H^{(L)} \right),$$
 with  $1 \le k \le K$. Similar to the Deep Scene-Sets architecture, $\phi^k$ are neural network modules with output vector length $D$ and parameter sharing in the last layer. To create a fixed vector representation, we combine all node features by the \text{sum} into an encoded \text{scene}. The Q-values can be computed by $Q_\mathcal{D} = Q(\Psi(X^{\text{dyn}_1}, ..., X^{\text{dyn}_K}) || x^{\text{static}})$. This module can replace the DeepScene-Sets module in DeepScene-Q as shown in Algorithm \ref{alg:sceneq} (Option 2) and in \Cref{fig:deepscene} (b).

\begin{algorithm}[t]
    \label{alg:sceneq}
\small
    \SetAlgoLined
    \DontPrintSemicolon
    initialize $Q_{\mathcal{D}}=(\phi^1, ...,  \phi^K,\rho, H, Q)$ and $Q'_{\mathcal{D}}=({\phi^1}', ..., {\phi^K}', \rho', H', Q')$, set replay buffer $\mathcal{R}$\\
    \For{\text{optimization step} o=1,2,\dots}{
     \SetKwProg{Fn}{}{:}{}
    
        get minibatch $(s_i,a_i,(X^{\text{dyn}_1}_{i+1}, ..., X^{\text{dyn}_K}_{i+1}, x_{i+1}^{\text{static}}),r_{i+1})$ from $\mathcal{R}$\\
        \ForEach{\text{transition}}{
          \ForEach{\text{object type} $k \in (1, ..., K)$}{
          \ForEach{\text{object }$x_{i+1}^j$ in $X^{\text{dyn}_k}_{i+1}$}{
                ${({\phi^k_{i+1}}')}^j = {\phi^k}' \left(x_{i+1}^j\right)$\\
            } 
            }
            \vspace{0.2cm}
          \Fn{{Set (Option 1) }}{
            \hspace{-0.3cm}    get  $\rho'_{i+1}=  \rho'\left(\sum\limits_k \sum\limits_j({\phi^k_{i+1}}')^j\right)$ 
            }
      
        \Fn{{Graph (Option 2) }}{
            \hspace{-0.3cm} compute $H^{'(L)}_{i+1}$ by GCN with 
              $H^{'(0)}_{i+1} = [\Phi^1, ..., \Phi^K ]^\top \text{ and }  \Phi^k = [(\phi'_{i+1})^1, ..., ({\phi'}_{i+1})^\text{seq len}]$\;
             
             \hspace{-0.3cm}  get  $\rho'_{i+1}=  \rho'\left(\sum\limits_k \sum\limits_j H^{'(L)}_{i+1} \right)$\;
        
            }
            $y_i = r_{i+1}+\gamma\max_a Q'(\rho'_{i+1}, x_{i+1}^{\text{static}}, a)$\\
        }
        perform a gradient step on loss and  update target network as in Algorithm \ref{alg:graphq}. 
    }
    \caption{DeepScene-Q}
\end{algorithm}

\subsection{Graph Construction}

We propose two different strategies to construct bidirectional edge connections between vehicles for Graphs and Deep Scene-Graphs representations:
\vspace{-0.1cm}
\begin{enumerate}
    \item Close agent connections: Connect agent vehicle to its direct leader and follower in its own and the left and right neighboring lanes ($6 \cdot 2$ edges).
    \item All close vehicles connections: Connect all vehicles to their leader and follower in their own and the left and right lanes ($K \cdot 6 \cdot 2$ edges for $K$ surrounding vehicles).
\end{enumerate}

Edge weights are computed by the inverse absolute distance between two vehicles, as shown in \cite{DBLP:journals/corr/abs-1903-01254}. A fully-connected graph is avoided due to computational complexity.

\subsection{MDP Formulation}

The feature representations of the the surrounding cars and lanes are shown in \cref{sec:inputfeatures}. The action space $\mathcal{A}$ consists of a discrete set of three possible actions in lateral direction: \textit{keep lane}, \textit{left lane-change} and \textit{right lane-change}.
 Acceleration and collision avoidance are controlled by low-level controllers, that are fixed and not updated during training. Maintaining safe distance to the preceding vehicle is handled by an integrated safety module, as proposed in \cite{deeptraffic,Mirchevska2018HighlevelDM}. If the chosen lane-change action is not safe, the agent keeps the lane. The reward function 
$r: \mathcal{S}\times\mathcal{A}\mapsto \mathbb{R}$ is defined as:
$r(s, a) = 1 - \frac{|v_{\text{current}}(s)- v_{\text{desired}}(s)|}{v_{\text{desired}}(s)} - p_{\text{lc}}(a),$
where $v_{\text{current}}$ and $v_{\text{desired}}$ are the actual and desired velocity of the agent, $p_{\text{lc}}$ is a penalty for choosing a lane-change action and minimizing lane-changes for additional comfort.

\begin{table*}[t!]
    \centering
    \footnotesize
    \begin{tabular}{c c c c c c }
    \toprule
         Driver Type & maxSpeed & lcCooperative & accel/ decel & length & lcSpeedGain\\
         \midrule
        agent driver & 10 & - & 2.6/4.5 & 4.5 & - \\
        passenger drivers 1&  $\mathcal{U}(8,12)$  & $0.2$ & 2.6/4.5 & $\mathcal{U}(4, 5)$&  $\mathcal{U}(5, 10)$ \\
        passenger drivers  2 & $\mathcal{U}(5,9)$  & $1.0$ & 2.6/4.5 & $\mathcal{U}(4,5)$ &  $\mathcal{U}(5, 10)$  \\
        passenger drivers  3 &  $\mathcal{U}(3, 7)$ & $0.8$ & 2.6/4.5 &  $\mathcal{U}(4, 5)$ &  $\mathcal{U}(5, 10)$ \\
        truck drivers   &  $\mathcal{U}(2, 4)$& $0.4$ & 1.3 / 2.25 &  $\mathcal{U}(9.5, 14.5)$ & $ \mathcal{U}(0,3)$ \\ 
        motorcycle drivers   & $ \mathcal{U}(7, 11)$ & $0.2$ & 3.0/5.0 & $ \mathcal{U}(2,3)$ &  $\mathcal{U}(15, 20)$\\
        \bottomrule
    \end{tabular}
    \vspace{0.2cm}
    \caption{SUMO parameters for different driver types. In each scenario, trucks and motorcycles are sampled with $10\%$ and $5\%$ probability, passenger cars and their driver types are sampled uniformly for the remaining number of vehicles.}
    \label{tab:driver_types}
\end{table*}

\section{EXPERIMENTAL SETUP}

We use the open-source SUMO \cite{SUMO} traffic simulation to learn lane-change maneuvers.

\subsection{Scenarios}

\paragraph{Highway} To evaluate and show the advantages of Graph-Q, we use the $\SI{1000}{\metre}$ circular highway environment shown in \cite{DeepSetQ} with three continuous lanes and one object class (passenger cars). To train our agents, we used a dataset with 500.000 transitions.  

\begin{figure}[b]
    \centering
       \includegraphics[width=0.48\textwidth]{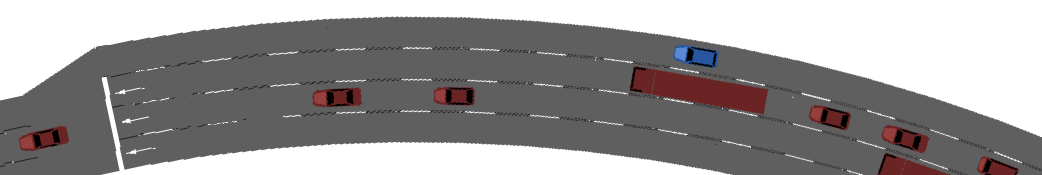}
    \caption{\textit{Fast Lanes} scenario in SUMO. The agent (blue) is overtaking other vehicles (red) on the fast lane and has to merge before the lane ends.}
    \vspace{-2mm}
    \label{fig:sumo}
\end{figure}

\paragraph{Fast Lanes} To evaluate the performance of DeepScene-Q, we use a more complex scenario with a variable number of lanes, shown in \Cref{fig:sumo}. It consists of a $1000\,$m circular highway with three continuous lanes and additional fast lanes in two $\SI{250}{\metre}$ sections. At the end of lanes, vehicles slow down and stop until they can merge into an ongoing lane. The agent receives information about additional lanes in form of traffic signs starting $\SI{200}{\metre}$ before every lane start or end. Further, different vehicle types with different behaviors are included, i.e. cars, trucks and motorcycles with different lengths and behaviors. For simplicity, we use the same feature representation for all vehicle classes. As dataset, we collected 500.000 transitions in the same manner as for the \textit{Highway} environment. 

\subsection{Input Features}
\label{sec:inputfeatures}
In the \textit{Highway} scenario, we use the same input features as proposed in \cite{DeepSetQ}. For the \textit{Fast Lanes} scenario, the input features used for vehicle $i$ are:

\begin{itemize}
\item \textit{relative distance}:
$dr_i = (p_i - p_{\text{agent}})/d_{\text{max}}\in\mathbb{R}$,\\  $p_{\text{agent}}$, $p_i$ are longitudinal positions in a curvilinear coordinate system of the lane.
          
\item \textit{relative velocity}:  
$ dv_i =  (v_i-v_{\text{agent}}) /v_{\text{allowed}}$
        
\item \textit{relative lane index}: 
$dl_i = l_i - l_{\text{agent}} \in \mathbb{N}$,\\
where $l_i$, $l_{\text{agent}} $ are lane indices.
        
\item  \textit{vehicle length}:  
$\text{len}_i / 10.0$
\end{itemize}

\noindent The state representation for lane $j$ is:
\begin{itemize}
    \item  \textit{lane start and end}: distances (km) to lane start and end
    \item   \textit{lane valid}: lane currently passable
    \item \textit{relative lane index}:  $dl_j = l_j - l_{\text{agent}} \in \mathbb{N}$,\\
    where $l_j$, $l_{\text{agent}} $ are lane indices.
\end{itemize}
For the agent, the normalized velocity $v_{\text{current}} / v_{\text{desired}}$ is included, where $v_{\text{current}}$ and $v_{\text{desired}}$ are the current and desired velocity of the agent. Passenger cars, trucks and motorcycles use the same feature representation. When the agent reaches a traffic sign indicating a starting (ending) lane, the lane features get updated until the start (end) of the lane.
\label{sec:appendix:states}

\subsection{Training \& Evaluation Setup}
All agents are trained off-policy on datasets collected by a rule-based agent with enabled SUMO safety module integrated, performing random lane changes to the left or right whenever possible. For training, traffic scenarios with a random number of  $n \in (30, 60)$ vehicles for \textit{Highway} and with $n \in (30, 90)$ vehicles for \textit{Fast Lanes} are used. Evaluation scenarios vary in the number of vehicles $n \in (30 , 35, ..., 90)$. For each fixed $n$, we evaluate 20 scenarios with different \textit{a priori} randomly sampled positions and driver types for each vehicle, to smooth the high variance. 

\label{sec:sumosettings}
In SUMO, we set the time step length to $\SI{0.5}{\second}$. The action step length of the reinforcement learning agents is  $\SI{2}{\second}$ and the lane change duration is $2\,s$. Desired time headway $\tau$  and minimum gap are  $\SI{0.5}{\second}$ and $\SI{2}{\metre}$. All vehicles have no desire to keep right ($\text{lcKeepRight} = 0.0$). The sensor range of the agent is $d_{\text{max}}=\SI{80}{\metre}$. \textit{LC2013} is used as lane-change controller for all other vehicles. To simulate traffic conditions as realistic as possible, different driver types are used with parameters shown in \Cref{tab:driver_types}.

\begin{table}[h!]
    \centering
    \footnotesize
    \begin{tabular}{c|c| c}
       \toprule
      Social CNN  & VBIN   & GCN\\
        \midrule
        \footnotesize{Input}($B \times 80 \times 5$) &  \footnotesize{Input}($B \times 15$) &  \footnotesize{Input}($B \times \text{seq} \times 3$)\\
        \midrule
         $\phi$: FC($20$), FC($80$) & $\phi$: FC($20$), FC($80$) & $\phi$: FC($20$), FC($80$)  \\
         $16 \times \text{Conv2D} (3 \times 1)$  & concat($\cdot$) & $1 \times \text{GCN}(80)$\\
         $32 \times \text{Conv2D} (3 \times 1)$ &   $\rho$: FC($80$), FC($20$) & sum($\cdot$)\\
        \midrule
         \multicolumn{3}{c}{concat($\cdot$, Input($B \times 3$))}\\ 
         \multicolumn{3}{c}{FC(100)$^*$, FC(100), Linear(3)}\\ 
         \bottomrule
    \end{tabular}
    \vspace{0.15cm}

     \begin{tabular}{c|c}
       \toprule
       Deep Scene-Sets & Deep Scene-Graphs\\
        \midrule
              \multicolumn{2}{c}{Input($B \times \text{seq}_0 \times 4$)  and Input($B \times \text{seq}_{1} \times 4$)} \\
        \midrule
         $\phi_{0}$: FC(20), FC(80),FC(80)$^{**}$ &  $\phi_{0}$: FC(20), FC(80),FC(80)$^{**}$\\
                 $\phi_{1}$: FC(20), FC(80), FC(80)$^{**}$&   $\phi_{1}$: FC(20), FC(80),FC(80)$^{**}$ \\
        sum($\cdot$) & $1 \times  \text{GCN}(80)$\\
        $\rho$: FC($80$), FC($80$) & sum($\cdot$)\\
        \midrule
         \multicolumn{2}{c}{concat($\cdot$, Input($B \times 3$))}\\ 
         \multicolumn{2}{c}{FC(100), FC(100), Linear(3)}\\ 
         \bottomrule
    \end{tabular}
    
        \caption{Network architectures. FC($\cdot$) are fully-connected layers. The CNN uses strides of $(2 \times 1)$. (*) For VBIN FC(200). (**) Parameters of the last layers are shared.}
        \label{tab:networks1}
\end{table}

\subsection{Comparative Analysis}

Each network is trained with a batch size of $64$ and optimized by Adam \cite{DBLP:journals/corr/KingmaB14} with a learning rate of $10^{-4}$. As activation function, we use Rectified Linear Units (ReLu) in all hidden layers of all architectures.  The target networks are updated with a step-size of $\tau=10^{-4}$.
All network architectures, including the baselines, were optimized using Random Search with the same budget of 20 training runs. We preferred Random Search over Grid Search, since it has been shown to result in better performance using budgets in this range \cite{Bergstra:2012:RSH:2188385.2188395}. The Deep Sets architecture and hyperparameter-optimized settings for all encoder networks are used from \cite{DeepSetQ}. The network architectures are shown in \Cref{tab:networks1}. Graph-Q is compared to two other interaction-aware Q-learning algorithms, that use input modules originally proposed for supervised vehicle trajectory prediction. 
To support our architecture choices for the Deep Scene-Sets, we compare to a modification with separate $\rho$ networks.
We use the following baselines\footnote{Since we do not focus on including temporal context, we adapt recurrent layers to fully-connected layers in all baselines.}:

\paragraph{Rule-Based Controller}
Naive, rule-based agent controller, that uses the SUMO lane change model \textit{LC2013}.

\paragraph{Convolutional Social Pooling (SocialCNN)}  In \cite{DBLP:journals/corr/abs-1805-06771}, a social tensor is created by learning latent vectors of all cars by an encoder network and projecting them to a grid map in order to learn spatial dependencies. 

\paragraph{Vehicle Behaviour Interaction Networks (VBIN)} In \cite{DBLP:journals/corr/abs-1903-00848},
instead of summarizing the output vectors as in the Deep Sets approach, the vectors are concatenated, which results in a limitation to a fixed number of cars. We consider the 6 vehicles surrounding the agent (leader and follower on own, left and right lane).

\paragraph{Multiple $\rho$-networks}
Deep Scene architecture where all object types are processed separately by using $K$ different $\rho$-network modules. The $K$ resulting output vectors are concatenated as 
$\big \lbrack \rho^1 \left( \sum_{x \in X^{ \text{dyn}_1}}\phi^1(x) \right), ...,  \rho^K \left( \sum_{x \in X^{ \text{dyn}_K}}\phi^K(x) \right) \big \rbrack$ and fed into the Q-network module.

\subsection{Implementation Details \& Hyperparameter Optimization}
\label{sec:appendix:networks}

All networks were trained for $1.25 \cdot 10^6$ optimization steps. The Random Search configuration space is shown in \Cref{tab:hyperopt}. For all approaches except VBIN, we used the same $\phi$ and $Q$ architectures. Due to stability issues, adapted these parameters for VBIN.  For SocialCNN, we used the optimized grid from \cite{DeepSetQ} with a  size of $80 \times 5$. The GCN architectures were implemented using the pytorch gemoetric library \cite{DBLP:journals/corr/abs-1903-02428}.

 \begin{table}[t!]
    \centering
    \begin{tabular}{c c c}
    \toprule
    Architecture & Parameter & Configuration Space\\
    \midrule
    Encoders & $\phi$: num layers &  $1,2,3$\\
             & $\phi$: hidden/ output dims & $5, 20, 80, 100$ \\
    Deep Sets & $\rho$: num layers & $1,2,3$\\
             & $\rho$: hidden/ output dims   & $5, 20, 100$\\
GCN & num  GCN layers & 1,2,3\\
& hidden  and output dim & 20, 80\\
& use edge weights & True, False\\
  SocialCNN & CONV: num layers & $2, 3$ \\
   &  kernel sizes& $([7, 3, 2], [2, 1])$  \\
   &  strides & $([2, 1], [2, 1])$  \\
    & filters & $8, 16, 32$  \\
    VBIN & $\phi$ : output dim & 20, 80 \\
    & $\rho$ : hidden dim  & 20, 80, 160, 200 \\
    & $Q$ : hidden dim & 100, 200\\
    Deep Scene-Sets & $\rho$ : output dim & 20, 80\\
             & shared parameters & True, False\\
             
    Deep Scene-Graphs & use $\rho$ network & True, False\\
           & $\rho$ : output dim & 20, 80\\
             & shared parameters & True, False\\
       \bottomrule
    \end{tabular}   
    \vspace{0.2cm}
    \caption{Random Search configuration space. For every architecture, we sampled 20 configurations to find the best setting.}
    \label{tab:hyperopt}
\end{table}

\begin{figure*}[t]
    \centering
       \includegraphics[width=0.39\textwidth]{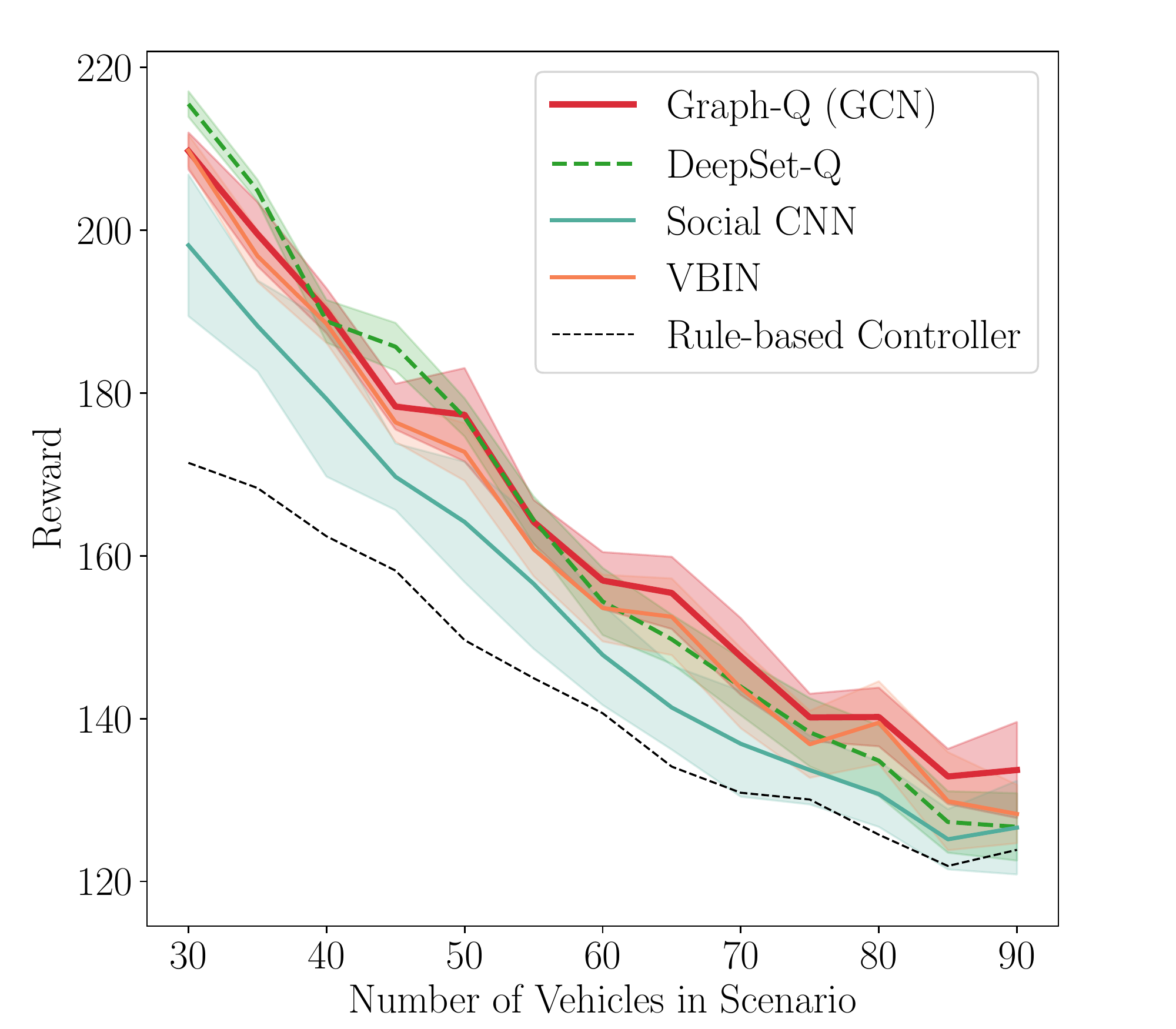}
       \includegraphics[width=0.39\textwidth]{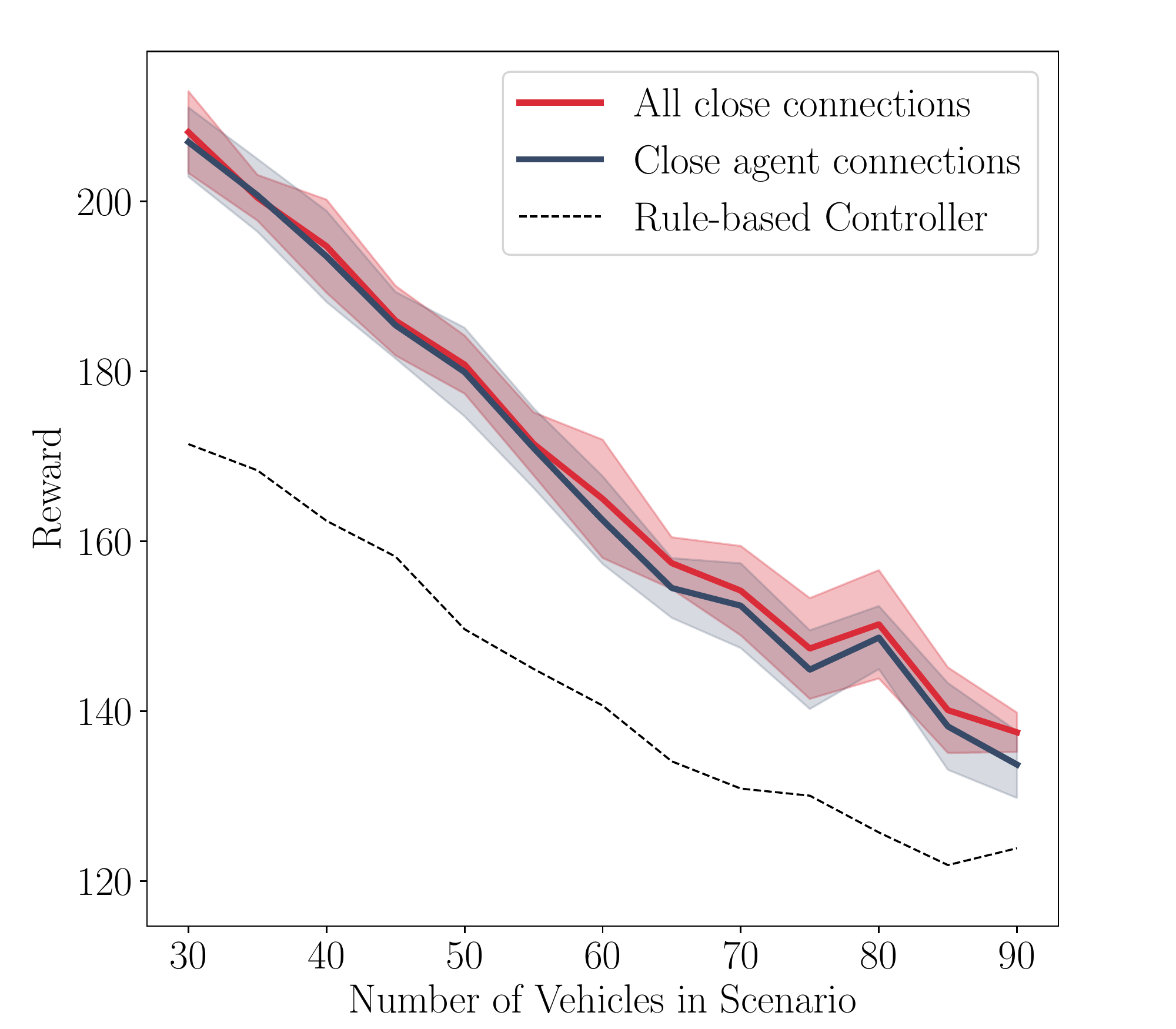}
\caption{ Mean performance and standard deviation in the \textit{Highway} scenario  over 10 training runs for Graph-Q with all close vehicle connections, the Deep Sets \cite{DeepSetQ} and two other interaction-aware Q-function input modules (left), and Graph-Q using the two proposed graph construction strategies (right). The number of vehicles indicates the traffic intensity, from light to dense traffic.}
\label{fig:perfhighway}
\end{figure*}

\begin{figure*}
    \centering
     \includegraphics[width=0.38\textwidth]{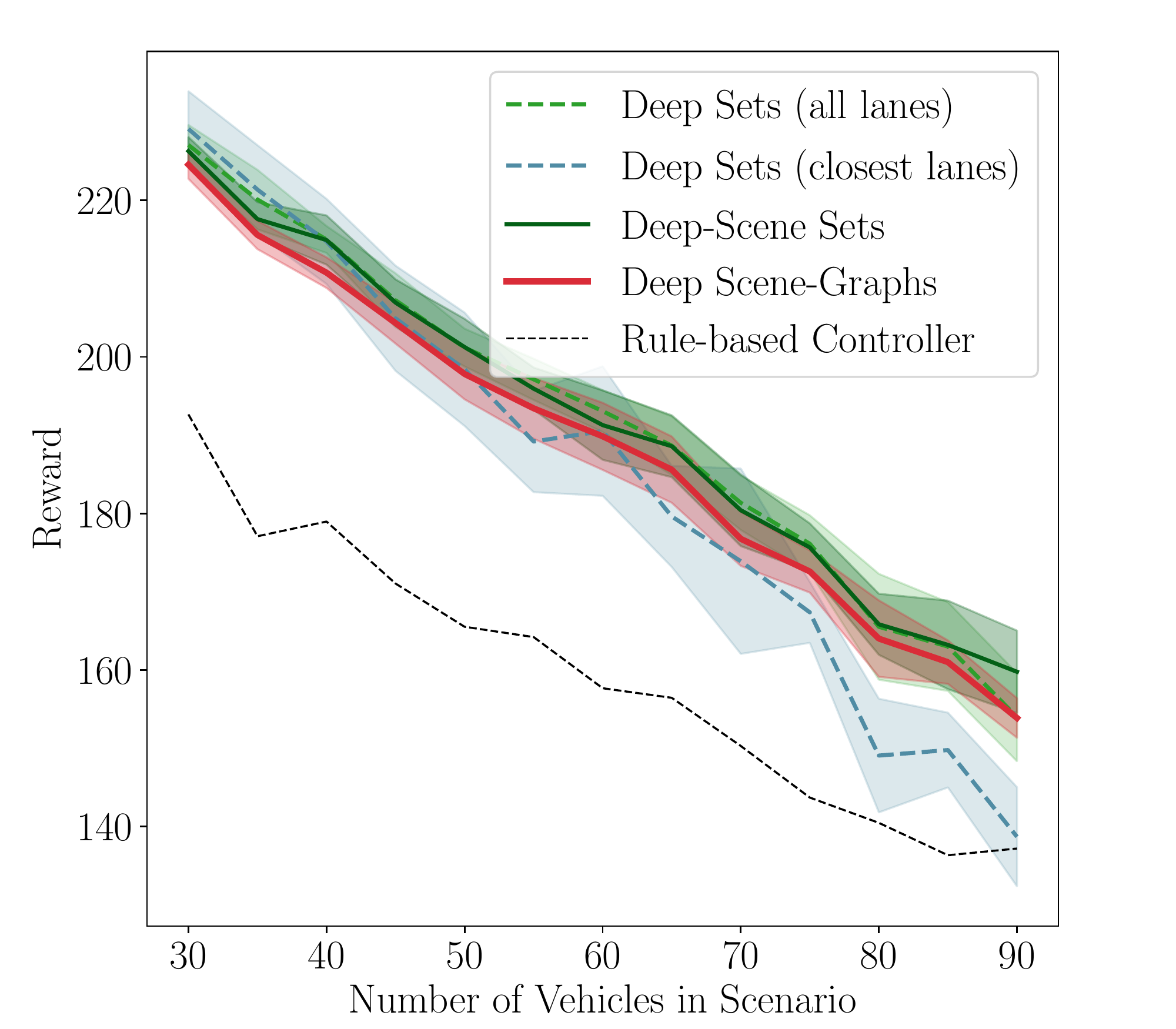}
     \includegraphics[width=0.38\textwidth]{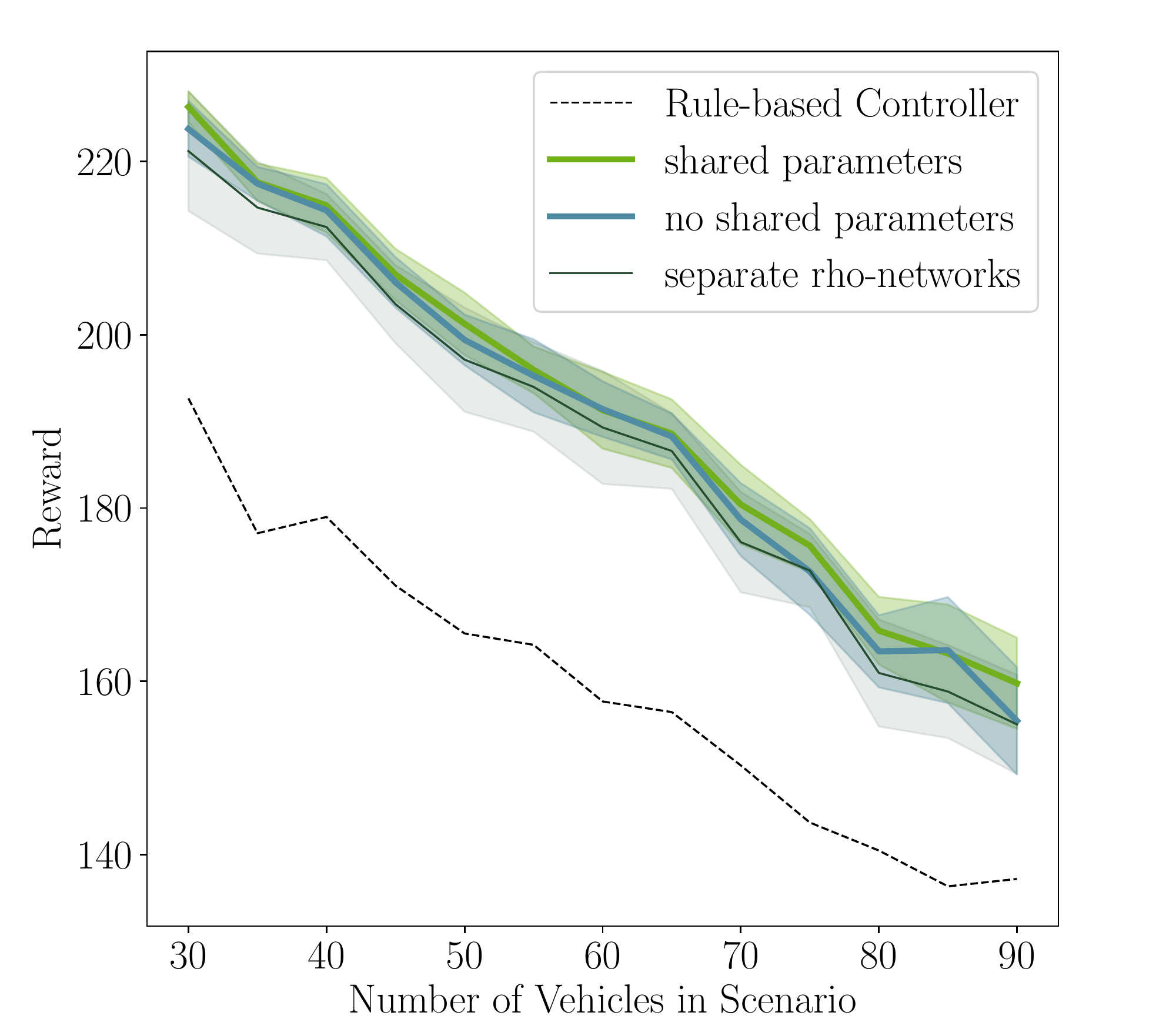}
\caption{Mean performance and standard deviation in the \textit{Fast Lanes} scenario over 10 training runs for Deep Scene-Sets, Deep Scene-Graphs and the rule-based controller from SUMO  (left), and different architecture choices of the Deep Scenes (right). The number of vehicles indicates the traffic intensity.}
\label{fig:perfdeepsetq}
\end{figure*}

\section{RESULTS}

The results for the \textit{Highway} scenario are shown in \Cref{fig:perfhighway}. Graph-Q  using the GCN input representation (with all close vehicle connections) is outperforming VBIN and Social CNN. Further, the GCN input module yields a better performance compared to Deep Sets in all scenarios besides in very light traffic with rare interactions between vehicles. While the Social CNN architecture has a high variance, VBIN shows a better and more robust performance and is also outperforming the Deep Sets architecture in high traffic scenarios. This underlines the importance of interaction-aware network modules for autonomous driving, especially in urban scenarios. However, VBIN are still limited to fixed-sized input and additional gains can be achieved by combining both variable input and interaction-aware methods as in Graph Networks.  To verify that the shown performance increases are significant, we performed a T-Test exemplarily for 90 car scenarios:
\begin{itemize}
    
\item  Independence of the mean performances of DeepSet-Q and Graph-Q is highly significant ($< 0.001$) with a p-value of 0.0011.
\item  Independence of the mean performances between Graph-Q and VBIN is significant ($< 0.1$) with a p-value of 0.0848. Graph-Q is additionally more flexible and can consider a variable number of surrounding vehicles.
\end{itemize}

\Cref{fig:perfhighway} (right) shows the performance of the two graph construction strategies. A graph built with connections for all close vehicles outperforms a graph built with close agent connections only. However, the performance increase is only slight, which indicates that interactions with the direct neighbors of the agent are most important.

The evaluation results for \textit{Fast Lanes} are shown in \Cref{fig:perfdeepsetq} (left). The vehicles controlled by the rule-based controller rarely use the fast lane. In contrast, our agent learns to drive on the fast lane as much as possible ($39.0\%$  of the driving time). We assume, that the Deep Scene-Sets are outperforming Deep Scene-Graphs slightly, because the agent has to deal with less interactions than in the \textit{Highway} scenario. Finally, we compare Deep Scene-Sets to a basic Deep Sets architecture with a fixed feature representation. Using the exact  same lane features (if necessary filled with dummy values), both architectures show similar performance. However the performance collapse for the Deep Sets agent considering only its own, left and right lane shows, that the ability to deal with an arbitrary number of lanes (or other object types) can be very important in certain situations. Due to its limited lane representation, the Deep Sets (closest lanes) agent is not able to see the fast lane and thus significantly slower. \Cref{fig:perfdeepsetq} (right) shows an ablation study, comparing the performance of the Deep-Scene Sets with and without shared parameters in the last layer of the encoder networks. Using shared parameters in the last layer leads to a slight increase in robustness and performance, and outperforms the architecture with separate $\rho$ networks.

\section{CONCLUSION} 
\label{sec:conclusion}

In this paper, we propose Graph-Q and DeepScene-Q, interaction-aware reinforcement learning algorithms that can deal with variable input sizes and multiple object types in the problem of high-level decision making for autonomous driving. We showed, that interaction-aware neural networks, and among them especially GCNs, can boost the performance in dense traffic situations. The Deep Scene architecture overcomes the limitation of fixed-sized inputs and can deal with multiple object types by projecting them into the same encoded object space. The ability of dealing with objects of different types is necessary especially in urban environments. In the future, this approach could be extended by devising algorithms that adapt the graph structure of GCNs dynamically to adapt to the current traffic conditions. Based on our results, it would be promising to omit graph edges in light traffic, essentially falling back to the Deep Sets approach, while it is beneficial to model more interactions with increasing traffic density.

\bibliographystyle{IEEEtran}
\clearpage

\balance
\bibliography{root}

\end{document}